\documentclass[runningheads,orivec]{llncs}
\usepackage[T1]{fontenc}
\usepackage{amsmath,amssymb}
\usepackage{graphicx}
\usepackage{todonotes}
\setuptodonotes{inline}
\usepackage{hyperref}
\usepackage{color}

\begin{document}
\title{An Affective-Taxis Hypothesis for Alignment and Interpretability}
%
%
\author{Eli Sennesh\inst{1}\orcidID{0000-0001-7014-8471} \and
Maxwell Ramstead\inst{2}\orcidID{0000-0002-1477-8177}}

\authorrunning{E. Sennesh and M. Ramstead}

\institute{Vanderbilt University, Nashville TN 37212, USA; \email{eli.sennesh@vanderbilt.edu} \and
Noumenal Labs and Queen Square Institute of Neurology, University College London;
\email{maxwell.d.ramstead@gmail.com}}
\maketitle              
\begin{abstract}
AI alignment is a field of research that aims to develop methods to ensure that agents always behave in a manner aligned with (i.e.~consistently with) the goals and values of their human operators, no matter their level of capability. This paper proposes an affectivist approach to the alignment problem, re-framing the concepts of goals and values in terms of affective taxis, and explaining the emergence of affective valence by appealing to recent work in evolutionary-developmental and computational neuroscience. We review the state of the art and, building on this work, we propose a computational model of affect based on taxis navigation. We discuss evidence in a tractable model organism that our model reflects aspects of biological taxis navigation. We conclude with a discussion of the role of affective taxis in AI alignment. 

\end{abstract}

\vspace{-1em}
\section{Introduction}
\label{sec:intro}
\vspace{-1em}

Artificial agents have recently taken up a prominent place in the field of artificial intelligence (AI). Technical progress has been impressive of late, with agents gradually performing more and more complex tasks, based largely on linguistic instructions. However, the state of the art machine learning techniques that are used to train these agents come with no guarantee that the decisions made and actions taken by agents designed using such methods always remain aligned (or consistent) with the goals and values of their human operators. AI \emph{alignment} focuses on ensuring that AI agents \emph{do} behave consistently in this way, even in the absence of direct human supervision.

Conventionally, since state of the art agents are trained using reinforcement learning, and accordingly optimize a reward, utility, or cost function over the state trajectories determined by their sequential decisions, AI alignment has focused largely on designing agents that share a reward (or utility, or cost, etc) function~\cite{Chichilnisky1985} with their human operators. Unfortunately, there are core obstacles that need to be overcome for this approach to get off the ground~\cite{Ramstead2025}. As pointed out in early work on AI alignment~\cite{Yudkowsky2004} and related philosophy literature~\cite{Railton1986}, while people's actions depend on the convolution of their values and their beliefs, their values themselves depend solely on the ground-truth state of affairs. This introduces interpretive ambiguity that potentially prevents explicit alignment. For example, depending on our beliefs, we might experience an elevated heart-rate~\cite{White1981} as either romantic attraction (positive, rewarding, valued) or as an effect of physical exercise (negative, costly, valueless). What we ``really'' value, in the sense necessary for building aligned autonomous agents, cannot be estimated solely from behavior, because it depends on a combination of how behavioral outcomes impact our physical reward systems and how those reward systems measure the cognitive costs of updating our beliefs along the way.

Mathematically, it is not possible to estimate the reward function of an agent based solely on observed behavior, because behavior is the convolution of beliefs and values. Without independently knowing about the belief formation mechanism at play, it is not possible to estimate the reward function used by an agent. (While we do not presently know what is the exact belief formation mechanism that is implemented by the human brain; whatever it is, however, we can be almost certain that it is not via the backpropagation of errors~\cite{Lillicrap2020,Song2024}).

State of the art approaches in machine learning, in some sense, have elided the problem of AI alignment altogether. Behavior cloning~\cite{Foster2024} is useful in practice because it does not require reward functions, but results in brittle AI systems that generalize poorly outside the training set~\cite{silver2025welcome}. Similarly, reinforcement learning from human feedback (RLHF~\cite{Bai2022}) does very well in many applied settings, but outsources the evaluation of system outputs (and therefore, the task of evaluating whether these outputs are aligned with human goals and values) to human raters. This does little but replace the reward function altogether with human raters. Ratings may be appropriate to train models that need to be deployed in specific, narrowly defined use cases (i.e.~those represented in the training data), but it does not get us any closer to achieving AI alignment. Neither does learning rewards from users by directly querying them about their preferences address the AI alignment problem, since we know from experimental psychology that self-report is a poor measure of actual preference~\cite{beshears2008preferences}. So what now?

We suggest that progress can be made here by explicitly considering human evaluative systems and designing artificial agents that conform to the same principles. Over the past few decades, the fields of psychology and neuroscience have fundamentally reconsidered the nature of evaluative processes in the human brain~\cite{Gundem2022,Weber2024}. Rather than assuming that there exist separate ``rational'' and ``emotional'' evaluative systems, often assumed to correspond to model-based and model-free neural processes, experimental evidence suggests that affect \emph{is} the core evaluative process~\cite{Railton2017,Dukes2021}, ubiquitous throughout all whole-brain activity~\cite{Barrett2009}. We have also learned that the sensorimotor grounding for affect rests in interoception~\cite{Barrett2017,Feldman2024}; we come to our decisions and select our actions on the basis of interoceptive beliefs, accountable to viscerosensory signals. 

Here we propose that if we want AI agents to align with us, then it ought to follow that AI agents must model both the formation of affective states in humans and the interoceptive facts that ground those states. This proposed grounding may provide an inductive bias to overcome the negative identifiability results in alignment approaches such as cooperative inverse reinforcement learning~\cite{HadfieldMenell2016,Armstrong2018}.

Accordingly, this paper proposes an ``affectivist''~\cite{Dukes2021} approach to AI alignment. Section~\ref{sec:affective_landscape} will start from the folk psychological intuitions behind the approach and review the psychological and neuroscientific arguments behind the approach, traveling back through evolutionary history to find a suitable grounding. Section~\ref{sec:open_issues} will then review progress and open problems in computational modeling along the lines of our approach. Section~\ref{sec:celegans} will review how the approach described can be applied to a well-studied model organism. We conclude in Section~\ref{sec:discussion} with discussions of limitations and directions for future research.

\vspace{-1em}
\section{Across the Affective Landscape}
\label{sec:affective_landscape}
\vspace{-1em}

Folk psychology and ordinary everyday experience suggest that we approach pleasant things and avoid unpleasant things. Within the affective sciences, the \emph{affective gradient} hypothesis~\cite{Shenhav2024} suggests that gradients over a landscape of affective hills and valleys can ``point the way'' for all motivated behavior, formalizing the folk-psychological intuition. As a brain basis of this hypothesis, the dopaminergic midbrain in vertebrates has long been studied as a brain basis for reinforcement learning, but such reinforcement learning (RL) studies tend to assume that ``reward'' (the psychological or mathematical construct) corresponds linearly to \emph{reward} (the physical \emph{stuff} given to an animal or participant in a study). While alternative models of dopaminergic midbrain have appeared in the literature, RL formalizes the expected utility metaphor, which lacks interpretability because it systematically identifies mathematical functions of distal or sensory states with physical substances (e.g.~fruit juice used to train an animal). No living organism whatsoever, no matter how simple, simply ``maximizes reward'' in the sense of collecting the largest amount of physical substance possible (whether water, fruit juice, food, etc) into its own body. The problem here is not mathematical models of the mind, but rather, their application to the study of motivated or goal-directed behavior without well-defined units.

Any interpretable model of a biological system's behavior that features well-defined units must account for evolutionary history and natural selection. In neuroscience and cognitive science, this project takes the form of phylogenetic refinement~\cite{Cisek2019}. That which was conserved through evolution, by powerful selection pressures preventing deviation, will tend to generalize across species, providing us with a historical ``inductive bias'' to understand that species-specific features emerged by elaboration and exaptation of what came before.

The evolutionary function of the first affective ``valence'' may have been to link an organism’s internal physiology with the movement of its body through space via taxis navigation~\cite{Bennett2021,Cisek2021,Barron2023}. While elementary taxis navigation emerged quite early, in single-celled organisms~\cite{Lazova2011,Kojadinovic2013} and was later replicated in multicellular animals, the earliest brains to feature taxis navigation likely evolved in early bilaterians~\cite{Bennett2021}. We propose that once taxis navigation evolved to navigate a bilaterian’s physical environment, the folding of the neural plate into a tube inside the organism reoriented the combined apical and blastoporal nervous systems towards navigating an internal taxis landscape (as described by Cisek~\cite{Cisek2021}), prescribing movement through a vector space of viscerosensory physiological indicators (i.e.~through interoceptive space~\cite{Keramati2014,Sennesh2021,Weber2024}). For reference purposes, call this the \emph{affective-taxis hypothesis}. We further suggest that the associative learning systems we typically consider the brain's ``reinforcement learning'' systems evolved later to invigorate or slow down taxis movement by estimating the long-run reward rate~\cite{Shadmehr2020}. This hypothesis could be checked neuroanatomically. In mammals, reward-taxis models~\cite{Karin2022} explain aspects of reward seeking in mice.

Unfortunately, in vertebrates such as mice and humans, associative reinforcement learning also ensures that behavior at any one moment does not reflect the immediate taxis landscape at that same moment; instead, it reflects the \emph{predicted, long-run} taxis landscape, given present sensory cues. To study taxis navigation as such rather than associative learning (i.e.~approximate dynamic programming), experimenters would either have to find a way to deactivate all learned associations while inhibiting their further learning, or study a model organism that simply lacks temporal reinforcement learning while nonetheless performing taxis navigation and sharing a common ancestor with vertebrates. 

This paper focuses its modeling work on the second alternative. The next section will review existing work on computational modeling of taxis navigation, while the section after that will consider a suitable model organism.

\vspace{-1em}
\section{Computational Modeling Progress}
\label{sec:open_issues}
\vspace{-1em}

Given a tractable experimental system, the above conceptual approach points to a number of immediate possibilities for technical progress in understanding evaluative reasoning in terms of affective gradients and reward-taxis. By considering ``reward'' as a spatial density over a physical or interoceptive space, the affective-taxis hypothesis falls into a common normative framework for modeling: gradient-biased random walks. Such gradient-biased random walks implement Bayesian inference over the unnormalized target density given by integrating the ``reward'' gradients over the whole space~\cite{Ma2015}. Accordingly, we begin by reviewing computational modeling approaches, particularly active inference approaches, which would explain the same gradient-biased random-walk behavior.

The literature on active inference has suggested other possible formalisms for affective valence, emphasizing the time-evolution or dynamics of the free energy. The free energy functional plays the same role in active inference as explicit reward functions in model-based reinforcement learning: namely, it provides an optimization objective. Initial work on this topic~\cite{joffily2013emotional} suggested linking affective valence to free-energy dynamics and learning rates: heuristically, when an agents' sensations fulfill its allostatic predictions, valence is positive and learning rate decreases, and vice-versa. Later work~\cite{hesp2021deeply} expanded on this idea to treat affective valence as the difference between expected and actual prediction error. Broadly, active inference provides a desirable modeling approach because it includes a normative standard for balancing exploration with exploitation~\cite{Keller1971,Keller1971b,Friston2015}.

In terms of the neural networks or other function approximators that would apply to a taxis navigation model, the affective-taxis hypothesis implies that we can model the unnormalized (log-)densities of the affective landscape with energy-based models~\cite{Song2021} (EBM's). EBM's enjoy compositional properties~\cite{Du2024} useful for passing from models of literal taxis in physical space to an internalized taxis in a constructed~\cite{Barrett2017,Westlin2023} and more abstract interoceptive space. Inspired by single neurons in the model organism that Section~\ref{sec:celegans} will introduce, we include as part of our affective-taxis hypothesis that scalar ``rewards'' capture the directional derivatives~\cite{Baydin2022,Fournier2023} of attractant gradients, enabling approximate gradient computation without backpropagation. Directional derivatives in EBM's could provide an alternative to reward languages or machines~\cite{Icarte2018}.

At the level of RL simulation environments, frameworks such as the Farama Gymnasium~\cite{Towers2024} could include taxis navigation or optimal foraging problems, including potentially by connecting with biophysical models of model organisms~\cite{Zhao2024}. ForageWorld~\cite{Badman2025} provides an example such environment, as do DeepMind's Rodent and Swimmer environments~\cite{Tunyasuvunakool2020}. Such an environment model would consist of a Partially Observable Markov Decision Process (POMDP, Definition~\ref{def:pomdp}), as is standard in RL.

\begin{definition}[Partially Observable Markov Decision Process]
\label{def:pomdp}
A \emph{Partially Observable Markov Decision Process (POMDP)} consists of a tuple
\begin{align*}
    (\mathcal{S}, \mathcal{A}, p_{S}, \mathcal{O}, p_{O}, R)
\end{align*}
whose components are a set of states $\mathcal{S}$, set of actions $\mathcal{A}$, probability density $p_S(s' \mid s, a)$ over state transitions, reward function $R: \mathcal{S} \times \mathcal{A} \rightarrow \mathbb{R}$, set of observations $\mathcal{O}$, and probability density $p_O(o \mid s)$ over observations given states.
\end{definition}
The unobserved state $s=(\mathbf{z}, \mathbf{v}, \gamma(\mathbf{z}; \beta), \beta)$ could consist of the model animal's location $\mathbf{z}$ and orientation $\mathbf{v}$ in allocentric coordinates, the spatial densities and placements $\gamma(\mathbf{z}; \beta)$ of attractants, and the physiological salience $\beta$ of each attractant. Actions $a=(\mathbf{a}, \alpha)$ could consist of linear and angular accelerations. Observations $o=(\nabla \log \gamma(\mathbf{z}; \beta))$ could represent attractant or repellent gradients at the current location, possibly perturbed by noise. Finally, rewards $R(\mathbf{s}, \mathbf{a})=\nabla_{\mathbf{z}} \log \gamma(\mathbf{z}; \beta) \cdot \mathbf{v}$ could consist precisely of the gradient's directional derivative with respect to the present velocity $\mathbf{v}$. Implementing this POMDP model will enable comparisons between controllers built by different methods such as simple RL, behavior cloning~\cite{Roberts2016,Chen2022,Barbulescu2023} and homeostatic RL~\cite{Keramati2014}.

The original run-and-tumble model for bacterial chemotaxis~\cite{Keller1971} showed how modulation of movement speed could balance exploitation and exploration in a field of attractants. Run-and-tumble does not capture the full range of taxis behavior in more complex species, and so ``solving'' the POMDP suggested above will require new normative models. These can begin by first extending the KL control model of \cite{Nakamura2022} beyond run-and-tumble, as well as integrating interoceptive incentive salience terms~\cite{Witham2016,Pool2016}. As taxis satisfies salient needs (e.g.~moving away from a noxious heat source), these salience terms will change, allowing another need to drive behavior; such gradient tempering~\cite{Karin2021,Zhu2018} can reproduce the Levy flights observed in foraging behavior~\cite{Moy2015}.

As the computational modeling efforts above become more sophisticated, we should also begin to study the inverse problem: how to observe taxis behavior and learn the time-varying ``surface'' over which that behavior steers. Such inverse RL learning of reward functions and behavioral policies is already a burgeoning field, although it does currently suffer from identifiability issues~\cite{Cao2021,Kim2021}. Energy-based models can capture such surfaces in an expressive, compositional way~\cite{Du2019}; training can proceed by forward gradients~\cite{Baydin2022,Fournier2023}, using observed movements as (negative) directional derivatives.

While the previous two sections have laid out the conceptual and computational framework for the affective-taxis hypothesis, they have avoided the temporal associative learning characteristic of standard RL. In RL terms, the temporal associative learning to which we refer would implement the ``critic'' state-value function in an actor-critic model of RL in the basal ganglia~\cite{Joel2002,Niv2009}. Studying the affective-taxis hypothesis in a real organism would thus require disentangling the present taxis landscape, whose temporal course the critic estimates, from the critic itself. To get around that issue, the next section will select and study a model organism for testing the hypothesis, a model for \emph{Bilateria} in which no such temporal associative learning occurs.

\vspace{-1em}
\section{A Model Organism's Affective Landscape: c. elegans}
\label{sec:celegans}
\vspace{-1em}

We argued above that AI alignment requires artificial agents to rigorously infer and represent affective states, specifically affective valence. The computational study of affect remains in its infancy, with most studies focusing either on applying theories of emotion to theory-of-mind tasks~\cite{Houlihan2023,Chen2024} or on the representation of emotion in AI models~\cite{Gandhi2024,Zhao2024emergence} -- rather than on formally modeling theories of how emotional and affective content arise in real brains. Here, we propose a computational model of a simple model organism engaging in whole-body behavior to regulate its internal physiology, and tie the biologically plausible components of that model to affect. This takes a first step towards understanding affect computationally such that AI engineers can rigorously design aligned agents.

Evolutionary approaches to any scientific question typically begin by generating and iterating on a hypothesis, and by testing it in a simple, experimentally tractable model organism. For bilaterians engaged in taxis via their central nervous systems, \emph{caenorhabditis elegans} serves as such a model organism~\cite{Larsch2015}. \emph{C. elegans} offers a number of advantages as a model organism, such as its sequenced genome and fully-mapped, stereotyped neuronal connectome. It also serves as a model organism for \emph{Bilateria}, going as far back in evolution as possible. For this problem, of understanding an organism's affective landscape, \emph{c. elegans} offers particular advantages as well: a number of the interactions between its interoceptive and sensory neurons have been mapped, its taxis behavior has stereotyped forms with tractable behavioral assays, and (most importantly for this purpose) a relative absence of associative reinforcement learning. While \emph{c. elegans} can learn to associate rewarding or punishing cues across sensory modalities, it does not associate them over time. Thus, taxis behavior in \emph{c. elegans} is driven by a present stimulus, not a prediction or belief about possible future stimuli, bypassing the need to infer such beliefs from behavior as in \cite{Ramstead2025}. This section will describe what is already known about taxis in \emph{c. elegans}.

Consider building the POMDP model proposed in Section~\ref{sec:open_issues} for \emph{c. elegans}. Observations could correspond to the signals transduced by taxis-related sensory~\cite{Lockery2011,Adler2014,Itskovits2018} and interoceptive~\cite{Witham2016,Hussey2018} neurons. The latter have been understood as reading off the internal state of the organism to provide salience weights~\cite{Hussey2018} for different ``rewards'' and ``costs'' outside the body. The former implement a transformation called \emph{fold-change detection}~\cite{Adler2014,Lang2016,Adler2018} (FCD). Colloquially, FCD signals a purely relative change in an input signal, rather than an absolute one; mathematically, FCD signals the time derivative of the logarithm of the input signal. In the case of sensory neurons for taxis, this input appears to be the spatial density of attractants and repellents. Denoting an animal's allocentric spatial location $\mathbf{z}(t)$ at time $t$ and its taxis-supporting sensory observations as $\mathbf{r}(t)$, FCD thus signals the log-density $\log \gamma(\mathbf{z}(t))$ of attractants and repellents

\begin{align*}
    \mathbf{r}(t) &:\approx \frac{d}{dt} \log \gamma(\mathbf{z}(t); \beta(t)),
\end{align*}
which equals the directional derivative of the attractant concentration with respect to the animal's present velocity
\begin{align*}
    \mathbf{r}(t) &= \nabla_{\mathbf{z}} \log \gamma(\mathbf{z}(t); \beta(t)) \cdot \frac{d\mathbf{z}(t)}{dt}, \\
    &= \|\nabla_{\mathbf{x}} \log \gamma(\mathbf{z}(t); \beta(t))\| \left\lVert \frac{d\mathbf{z}(t)}{dt} \right\rVert \cos{\theta},
\end{align*}
with $\theta$ the angle between the direction of movement and the spatial gradient. The spatial gradient provides a reference signal (in the sense of control theory) for taxis navigation. Seen another way, following the spatial gradient by turning to maximize $\mathbf{r}(t)$ and moving forward implements noisy gradient-based optimization equivalent to Langevin dynamics~\cite{Keller1971,Keller1971b}
\begin{align*}
    \frac{d\mathbf{z}(t)}{dt} &= \nabla_{\mathbf{z}} \log \gamma(\mathbf{z}(t); \beta(t)) + \sqrt{2}d\sigma(t).
\end{align*}

While direct experimental evidence remains limited, the hypothesis that certain sensory neurons implement fold-change detection may apply in vertebrate species as well~\cite{Adler2018}. Direct evidence that interoceptive or similar sensory neurons (such as olfactory neurons) implement FCD would support the affective-taxis hypothesis of how behavior serves to control internal physiology.

Above, we proposed that signaling the (negative) directional derivative of an energy function provides a reference signal for valence, one of the two dominant components of core affect~\cite{Barrett2009} and the computational basis for behavioral reinforcement learning. The other such component is arousal; studies have suggested examining wakeful, autonomic, and affective arousal~\cite{Satpute2019}. Unfortunately, computational models have tended to systematically conflate not only these different kinds of ``arousal'', but arousal itself with valence. We note that arousal is a complementary affective dimension to valence, with both energetic movement and behavioral quiescence sometimes taking place in situations of different intuitive valences (e.g.~fleeing a predator in an aroused state vs resting contentedly in a quiescent state). Most reinforcement learning theories of the human brain tend to associate arousal with the value of an action with the motor vigor of that action through the neuromodulator dopamine~\cite{Barron2023,Shadmehr2020}; some now tie the value-prediction aspect of RL to serotonin~\cite{Harkin2025}. Each of these neuropeptides play a variety of roles even in \emph{Bilateria} like \emph{c. elegans} with stereotyped connectomes.

RL theories of the human brain tend to associate dopamine with reward prediction error, the invigoration of movement, and subjective arousal~\cite{Barron2023,Shadmehr2020}. Active inference models of vertebrate behavior cast dopamine as a precision parameter measuring confidence in sensory cues about actions~\cite{Friston2012}. In \emph{c. elegans}, activating dopaminergic neurons makes worms slow down into a ``crawling'' gait and behave as if exploiting a patch of resources~\cite{VidalGadea2011}, while dopamine also serves an essential role in controlling the speed (vigor) of movement~\cite{Omura2012,Ji2023}, as in vertebrates. We suggest here that, from an evolutionary perspective, dopamine controls the gain of proprioceptive feedback from motor movements to enable an organism to exploit opportunities in its nearby environment.

Less is known as a whole about the neural role of serotonin. Theoretical work in animal reinforcement learning suggests that tonic serotonin might encode the infinite-horizon reward rate~\cite{Daw2002,Harkin2025}, but other work has complicated that view by associating serotonin with apparent behavioral quiescence or disengagement. This perspective is supported by evidence from \emph{c. elegans}: serotonergic activation makes worms change their locomotory gait, as if leaving patches of resources~\cite{VidalGadea2011}. The sensory endings of the neurosecretory motor neurons located in the pharynx release serotonin when they sense food, leading to locomotor slowing, pharyngeal pumping, and eventually egg-laying behaviors~\cite{Chase2007}. We suggest here that serotonin signals a shift from using the body to exploit environmental ``rewards'' to spending energy on the internal physiological processes requiring those ``rewards'', tempering~\cite{Karin2021} the taxis landscape to inhibit locomotion. 

We leave the further modeling of arousal to future work, while noting that its neural bases may involve tempering~\cite{Karin2021} of the taxis gradients described above.

\vspace{-1em}
\section{Discussion}
\label{sec:discussion}
\vspace{-1em}

This paper proposed to further the discussion of AI alignment by considering the fact that human beings are affective agents with affective states. We proposed that an aligned AI must accordingly be able to explicitly represent affective states, and in particular, affective valence. We thus proposed a computational account of the evolutionary roots of evaluative reasoning; our account dovetails with approaches to AI alignment in both the inverse reinforcement learning and active inference literature. In our view, affective-taxis (alongside other computational accounts of affect~\cite{Emanuel2023,hesp2021deeply,Shenhav2024}) provides a key contribution to alignment because affect underlies evaluative reason~\cite{Railton2017}. It follows that an aligned AI agent must necessarily represent people's affective states and beliefs.

\paragraph{Limitations} We have only considered a computational model of a very simple model organism with a rudimentary valence response; our model organism in this paper cannot even engage in associative learning across time. An affective-taxis model for humans would require considering far more dimensions than simple metabolic or thermoregulatory states. It crucially requires considering the associations across time, situated conceptualizations, and theory of mind that the human brain constructs as part of its affective states.
Future work should both test the computational model discussed above in simple organisms, and extend that model down the human lineage to capture evaluative phenomena that may have evolved~\cite{Bennett2021} in vertebrates (temporal associative learning), mammals (embodied simulation and model-based reasoning), and primates (theory of mind).

\paragraph{Related Work} The hypotheses and research program described in this paper intersect with the emerging subfield of ``NeuroAI'' and its applications to AI alignment. In particular, this paper suggests that the ``reward function in the brain'', as discussed by \cite{Mineault2024} and \cite{Byrnes2025}, consists of the (negative) directional derivative, in the present direction of movement, of a time-varying, interoceptive energy function; note that ``energy'' here refers to a function $E(\mathbf{z}) = -\log \gamma(\mathbf{z})$, as in energy-based models, rather than to physical energy. The time-varying, interoceptive nature of this function ties it to the physiological theory of allostasis~\cite{Sterling2012,Sennesh2021} (``stability through change'') and the allostatic roots of affect and emotion~\cite{Barrett2017,Feldman2024}. While any reward function can be made Markov by adding dimensions to the underlying state and action spaces over which it is defined~\cite{Abel2021}, by default a Markov reward function over states of an organism's environment, rather than the organism's internal states, cannot capture affect, and therefore the brain's ``reward functions'', in the way that allostatic energy functions can. Thus, allostatic energy functions capture ``time-extended preferences like the desire to keep a promise, or the value of narrative coherence''~\cite{ZhiXuan2024} while also providing compositional semantics~\cite{Du2024}.

\paragraph{Summary} Building on recent work in affective science~\cite{Shenhav2024} and computational modeling~\cite{Karin2022} this paper proposed an \emph{affective-taxis hypothesis} of motivated behavior and tied it to interpretability and alignment of AI systems. Section~\ref{sec:affective_landscape} described the core hypothesis across species, Section~\ref{sec:open_issues} reviewed progress in computational modeling of our hypothesis, and Section~\ref{sec:celegans} described its application to a model organism where evidence supports the hypothesis.

\begin{credits}
\subsubsection{\ackname} The authors thank Candice Pattisapu for insightful comments and editing recommendations, as well as Max Bennett for early discussions.

\subsubsection{\discintname}
The authors have no competing interests to declare that are relevant to the content of this article.
\end{credits}
%
%
%
%

\bibliographystyle{splncs04}
\bibliography{samplepaper.bib}

\end{document}